\documentclass[journal]{IEEEtran}

\usepackage[pagewise, switch]{lineno} 

\ifCLASSINFOpdf
\else
\fi
\usepackage{xcolor}

\usepackage[nocompress]{cite}
\usepackage{url}
\usepackage{amsmath,amssymb,graphicx}
\usepackage{multirow}
\usepackage{array}

\newcommand{\thickhline}{%
	\noalign {\ifnum 0=`}\fi \hrule height 1pt
	\futurelet \reserved@a \@xhline
}
\newcolumntype{"}{@{\hskip\tabcolsep\vrule width 1pt\hskip\tabcolsep}}

\begin{document}
%
\title{Human-Imperceptible Identification with Learnable Lensless Imaging}
%
%
%

\author{Thuong~Nguyen~Canh,
        Trung~Thanh~Ngo,
        and Hajime Nagahara
}

%
%

\markboth{Human-Imperceptible Identification with Learnable Lensless Imaging}%
{Shell \MakeLowercase{\textit{et al.}}: Bare Demo of IEEEtran.cls for IEEE Journals}
%



\maketitle

\begin{abstract}
Lensless imaging protects visual privacy by capturing heavily blurred images that are imperceptible for humans to recognize the subject but contain enough information for machines to infer information. Unfortunately, protecting visual privacy comes with a reduction in recognition accuracy and vice versa. We propose a learnable lensless imaging framework that protects visual privacy while maintaining recognition accuracy. To make captured images imperceptible to humans, we designed several loss functions based on total variation, invertibility, and the restricted isometry property. We studied the effect of privacy protection with blurriness on the identification of personal identity via a quantitative method based on a subjective evaluation. Moreover, we validate our simulation by implementing a hardware realization of lensless imaging with photo-lithographically printed masks.

\end{abstract}


%
\IEEEpeerreviewmaketitle

\section{Introduction}

\IEEEPARstart{V}{isual} privacy is the right to limit the collection and use of visual information related to personal identity. With the proliferation of cameras in smartphones, wearable devices, laptops, and surveillance equipment, it is increasingly easy to photograph someone without being noticed, violating their visual privacy. At the same time, visual appearance, such as facial data, has become the most commonly used feature in biometric authentication for accessing private information \cite{Lopez_Privacy_review,cavailaro_privacy_2007}, and its usage has significantly increased in recent years. Additionally, advanced virtual and augmented reality headsets are equipped with multiple cameras, which are most likely used in work and personal environments \cite{wang_wearable_2016} containing sensitive visual information. Alongside these hardware developments, artificial intelligence has advanced accuracy in many computer vision tasks, raising the question of privacy. In fact, recent research \cite{Montreal_AI} identified privacy protection as one of the most important ethical principles of artificial intelligence. Therefore, imaging systems should consider not only its functionality but also its ability to protect visual privacy. Fortunately, computer vision tasks can be performed without knowledge of subject identity \cite{winkler_privacy_2013, ReconsFreeVision_19, LowQualityVision}.

On the other hand, visual privacy can be classified as the protection of sensitive visual information against machine vision and/or human visual systems. For example, visual privacy against machine vision using adversarial samples, a sub-topic of adversarial machine learning \cite{kurakin2016adversarial}, aims to fool the targeted machine vision system \cite{yuan2019adversarial}. For example, adversarial images usually succeed in fooling machine recognition via the addition of noise, objects, or via cartoonization, but often fail to fool human vision \cite{wiyatno2019adversarial}. It should be noted that we pursue a completely different direction. Our goal is to deceive human vision by capturing a human-imperceptible image without preventing machine recognition. Therefore, we can enable computer vision applications without compromising visual privacy. 

In general, it is preferable to implement a high degree of visual protection at the physical or sensor level to prevent attacks during information transmission and to reduce the vulnerability to software hacks. One promising method is secure imaging. Secure imaging is an active research area that studies image capture using secure formats, such that the captured images are visually protected. Unlike many software-based approaches that are vulnerable to malware \cite{Winkler_TrustEye} \cite{shu_cardea_2018}, the output of an image sensor is secure and difficult to hack, thus enabling hardware-level privacy protection.
Previous studies have applied several methods for developing secure cameras \cite{Pittaluga_PrivacyOptic}, \cite{Pittaluga_PrivacyThermal}. Among them, lensless imaging \cite{LenslessCS} is the most common method. Lensless imaging processes incoming light by replacing the lens system with a coded mask \cite{LenslessCS} and reconstructs a visually understandable image from the captured measurements. Researchers have realized lensless imaging based on compressive sensing (CS) technology, developing systems such as a single-pixel camera with a directed micromirror \cite{SinglePixelCam}, a single-pixel lensless imaging system with a coded liquid crystal display (LCD) \cite{FlatCam}, \cite{boominathan2016lensless}. These systems capture a very blurry image, which represents a simple way of protecting visual privacy but paid no attention to the trade-off between visual privacy and recognition accuracy \cite{canopic}. 


In this work, we make the following contributions:
\begin{itemize}
    \item We propose an end-to-end framework for learning the coded pattern of a human-imperceptible recognizer;
    \item We propose similarity, total variation (TV), invertibility, and restricted isometry property  (RIP) losses to protect visual privacy against human vision;
    \item We evaluate visual privacy using the blurriness of captured measurements, a RIP score, and perceptual experiments on face identification;
    \item We implement a lensless imaging system with various printed masks.
\end{itemize}

The rest of this paper is organized as follows. We review related work in Section~\ref{sec:related_work}. Section~\ref{sec:secure_imaging} introduces our learnable lensless imaging system for recognition, and Section~\ref{sec:human_imperceptible_loss} presents various loss functions for preserving visual privacy. We describe the perceptual evaluation experiment in Section~\ref{sec:exp} alongside a hardware realization of our system, and we present our conclusions in Section~\ref{sec:conclusion}.

\section{Related Work} 
\label{sec:related_work}
To protect visual privacy, together with cartoonization, blanking, scrambling, blur the sensitive region is a common practice to protect visual privacy again human visual system in both academic \cite{Winkler_TrustEye}, \cite{Cambareri_PrivacyCS}, \cite{yu_iprivacy_2017} and industry (i.e. Google map, etc.). In general, encoder side protection is preferred to prevent transmission attach and could be further classified into the two approaches discussed below.

\subsection{Software-based Approach} Software-based approaches \cite{koelle_your_2018}, \cite{yu_iprivacy_2017} process the captured image by applying cartoonization, blanking, and scrambling sensitive regions \cite{Winkler_TrustEye}, \cite{Cambareri_PrivacyCS} before performing high-level computer vision tasks. However, private visual information remains available to software attacks \cite{Lopez_Privacy_review}, because the sensitive regions still exist in the image captured by a regular camera before processing. In addition, a single failure in detecting sensitive regions may reveal the identity of objects. CS theory has been adopted to capture secure images \cite{Calderbank_compressedlearning}, \cite{Cambareri_PrivacyCS}, \cite{Lohit_CNNInfer} with a stronger level of protection. That is the whole is mage is imperceptible, thus avoiding detection failure. Recent studies have shown that it is possible to understand the context from CS measurements with \cite{Huang_VQA_CS} or without reconstruction \cite{Calderbank_compressedlearning}, \cite{Lohit_CSInference}.

\subsection{Hardware-based Approach} For protection at the hardware level, researchers have used thermal images \cite{Pittaluga_PrivacyThermal}, \cite{zhang2014anonymous}, defocused images \cite{Pittaluga_PrivacyOptic}, and low-resolution images \cite{LowQualityVision}, \cite{canh2017privacy}. Tan et al. \cite{FlatCam_Face} achieved remarkable accuracy using the lensless FlatCam system \cite{FlatCam} for face detection and verification tasks. Canh et al. \cite{Canh_2019_ICCV} further developed FlatCam for visual privacy applications by protecting sensitive regions in the measurement domain. However, both methods require an initial reconstruction stage, which remains vulnerable to software attacks.
A previous study has investigated reconstruction-free inference for lensless action recognition \cite{FlatCam_Action}. The authors of this study captured multiple lensless measurements and used phase correlation and translation motion features to recognize action accuracy regardless of the coded patterns. Their method requires no reconstruction, thus avoiding the training complexity associated with learning multiple patterns. Tan et al. \cite{canopic} further enhanced visual privacy protection by considering additional analog processes such as quantization.
While previous work \cite{FlatCam_Action} has determined that measurements captured by lensless imaging are generally unrecognizable to humans, the accuracy tradeoff between human-based and machine-based recognition has received little or no attention \cite{canopic}. The extent to which lensless imaging impacts the human perception of identity remains unknown, with a lack of research on perceptual and objective quality assessment for recognition.


\subsection{Visual Privacy of Lensless Imaging}
In general, the blurriness of lensless coded images hides visual information, and therefore it has been assumed to protect visual privacy in lensless imaging \cite{FlatCam_Face, Canh_2019_ICCV}. 
Conversely, most results from the image quality assessment field have focused on the quality of the displayed image, which is not directly related to the recognition capability of blurry images. Previous studies \cite{Noref_Lightfield, Vinh} have evaluated the blurriness of images but have often considered motion blur and out-of-focus blur, which are not associated with recognition accuracy. Another concern is the reconstruction of a lensless image can reveal visual information via a blind deconvolution~\cite{Lam_00}. Therefore, reducing invertibility (i.e., the ability to reconstruct clear images from blurry images) is an essential factor in protecting visual privacy. To address this requirement, we use the restricted isometry property (RIP) \cite{candes_rip,canh_rsrm} to measure the invertibility of lensless images.

\begin{figure}[!t]
	\centering
	\includegraphics[scale=0.29]{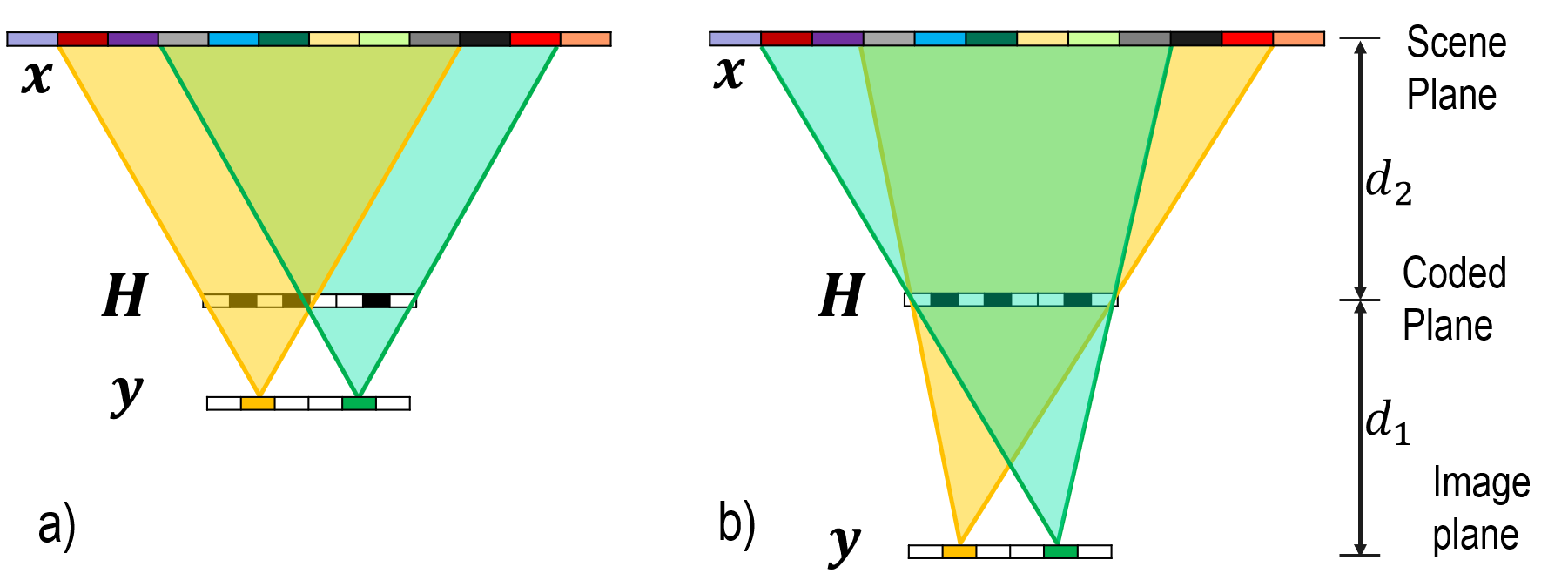}
	\scriptsize
	\caption{Lensless imaging with (a) short distance and (b) long distance between the image sensor and coded pattern.}
	\label{Fig_Lensless}
\end{figure}

Additionally, most previous work on visual privacy has used manually coded patterns to capture blurry images but achieved poor recognition accuracy. Although it is possible to learn an efficient pattern for high accuracy, this also reduces the blurriness of captured images and thereby reduces visual privacy. Moreover, the relationship between blurriness and visual privacy protection has not been addressed. Our work subjectively evaluates the recognition of lensless images and proposes a method to protect visual privacy while maintaining high recognition accuracy.
Researchers have simply associated the blurriness of coded measurements with visual privacy protection \cite{Canh_2019_ICCV, FlatCam_Face}. However, how well lensless imaging can protect the visual privacy of coded measurements for recognition is still unknown.

\section{Secure Lensless Imaging}
\label{sec:secure_imaging}

\subsection{Lensless Imaging}
Lensless imaging is an image capture technique without a complex lens system. A coded pattern is used to modulate the incoming light with single-pixel \cite{LenslessCS} or multiple-pixel \cite{FlatCam} cameras. As camera sensors have become less expensive, the latter approach has become more popular, because it enables single-shot image capture without changing coded patterns. As shown in Fig. \ref{Fig_Lensless} (a)--(b), given an image $x$ and a coded pattern $H$, the lensless measurement $y$ is modeled by
\begin{equation}
y = H * x + \eta,
\end{equation}
where  $*$ is the convolution operator and $\eta$ represents the additive noise source.
When the distance $d_1$ between the image and the corresponding coded pattern is decreased, the camera becomes thinner \cite{FlatCam}. This factor also limits the angle of the incoming light rays and thereby restricts the field of view. At the same resolution of $H$, a larger $d_1$ leads to a blurrier image because of the large kernel size, thereby providing better visual privacy. We, therefore, adopt a large $d_1$ value, and we ignore noise and diffraction for simplicity.

\begin{figure*}[!t]
	\centering		
	\includegraphics[scale=0.3]{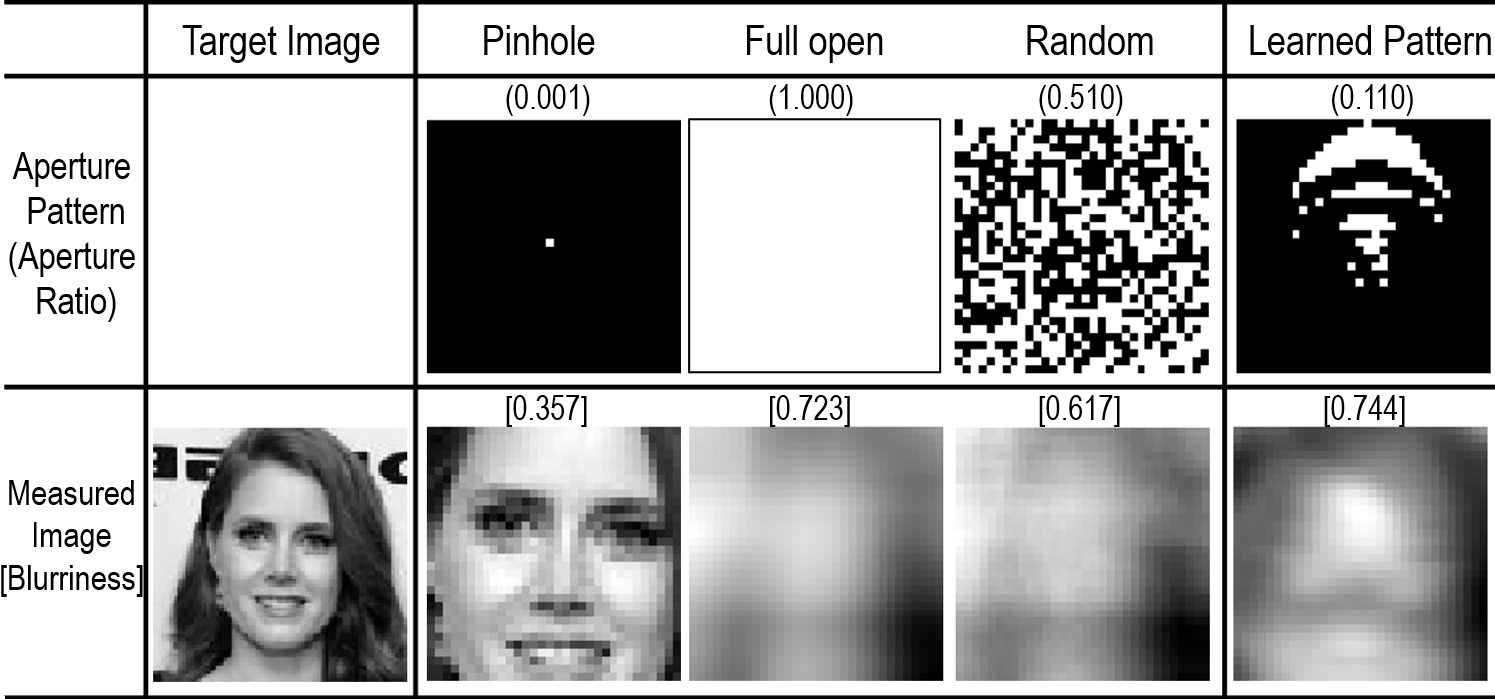}
	\caption{Visual comparison among different coded patterns with fixed (pinhole, full open, random) and learned pattern. Numerical values above measured images and corresponding patterns indicate [blurriness] \cite{crete2007blur} and (aperture ratio), respectively. The aperture ratio is the proportion of “bright” (equal to 1) pixels in the pattern. }
	\label{Fig_VisualFix1}
\end{figure*}

Fig. \ref{Fig_VisualFix1} shows examples of various fixed patterns (Pinhole, Full open and random), our proposed (Learned pattern), and their corresponding captured images. Subject identity is seen by the pinhole pattern, but not by the random and fully open patterns. Lacking an objective metric to evaluate visual privacy, previous work has relied on the blurriness metric as a simple privacy indicator \cite{crete2007blur}. Unfortunately, blurriness \cite{crete2007blur} is not a satisfactory measure of human imperceptibility: for example, higher blurriness nevertheless reveals more information in the learned pattern of Fig. \ref{Fig_VisualFix1}. There is a demand for a better evaluation method. In later work, we associate the visual privacy of lensless imaging with the ability to reconstruct an image from its measurement. Therefore, we proposed approximated inverse RIP property as a criterion in Section IV. 

\begin{figure}[!t]
	\centering
	\includegraphics[scale=0.31]{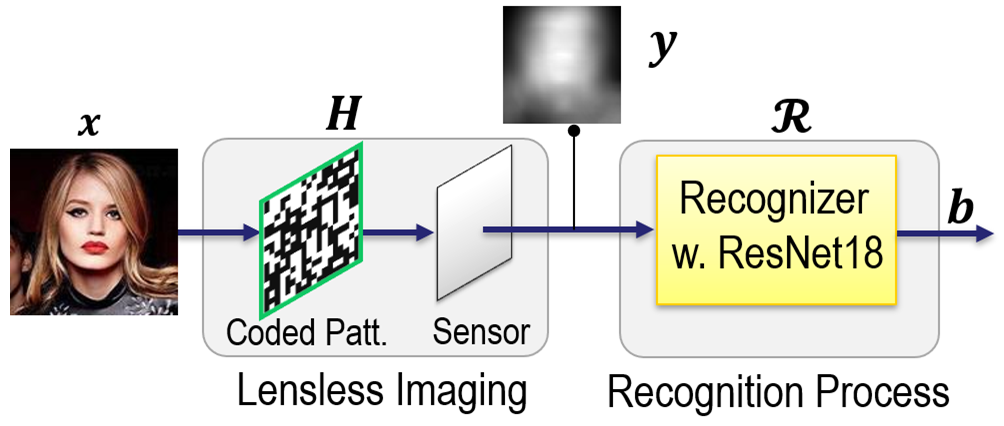}
	\scriptsize
	\caption{Proposed face recognition system with lensless imaging.}
	\label{Fig_FaceRegSys}
\end{figure}

\subsection{Lensless Imaging for Human-Imperceptible Identification}
Our lensless imaging system for identification, illustrated in Fig. \ref{Fig_FaceRegSys}, captures images using a lensless camera with a pattern $H$, and then sends them to a recognizer without reconstruction. An early study \cite{FlatCam_Action} used hand-crafted patterns, but the better performance was subsequently achieved with learned patterns \cite{iliadis2020deepbinarymask, Okawara}. Therefore, our approach involves joint learning of both the $H$ pattern and the recognizer, while preserving human imperceptibility and preventing reconstruction. It is straightforward to use the identification loss $\mathcal{L}_{\mathsf{rec} }(\mathcal{R}(H * x), b)$ for training this network, where $\mathcal{L}_{\mathsf{rec}}$ denotes the cross-entropy loss with a recognizer $\mathcal{R}$, an input image $x$, and a label $b$. The total training loss $\mathcal{L}$ is
\begin{equation}
\mathcal{L} = \mathcal{L}_{\mathsf{rec}}(\mathcal{R}(H * x), b) + \alpha \mathcal{L}_{\mathsf{hi}},
\end{equation}
where $\mathcal{L}_{\mathsf{hi}}$ denotes human-imperceptible loss (further detailed in the next section).
We adopted the method of \cite{iliadis2020deepbinarymask} to learn the binary patterns $H$, and we used ResNet18 as an example for the recognizer.

\subsection{Scope of Proposed Framework}

As the concept of privacy is very broad, only visual privacy is considered in this work. In traditional imaging, privacy is protected at the software level, which is vulnerable to software attacks. In the worst-case scenario, attackers can gain access to the camera output and retrieve all available privacy information. In contrast, lenselss imaging offers privacy protection at a physical level. Even in the event of completely successful attacks, attackers only gain access to the blurred measurement. 

Our proposed lensless framework is able to fool both human and machine vision to a certain extent. On one hand, the captured image is extremely blurry and imperceptible to the human visual system. On the other hand, without knowing the coded pattern, attacking the lensless measurement without knowing the coded pattern results in poor reconstruction. In order to guess the coded pattern, attackers must engage in a plain-attack scenario under which they gain the ability to (1) manipulate the input and output of the camera, (2) collect multiple images, and (3) perform training geared towards guessing the coded pattern.


\section{Human-Imperceptible Loss}
\label{sec:human_imperceptible_loss} 
This section describes loss functions that balance privacy and recognition performance. We propose four methods in the form of the loss function to learn the coded patterns. These methods are based on different components of the imaging model and are motivated by theoretical considerations. We first introduce measurement-based similarity loss, followed by loss based on coded patterns in combination with TV, invertibility, and RIP.

\subsection{Measurement-Based Human-Imperceptible Loss}

In general, the loss function should be some objective quality metric that is highly correlated with the results of the subjective evaluation described in Section \ref{sec:subjectTest} and Fig. \ref{fig_subjective_accuracies}. Unfortunately, there are no existing visual privacy metrics. To address this issue, we devised a protocol for assessing the impact of blurriness on human perception of blurred images. A simpler approach may involve maximization of the difference between input images and corresponding captured images, however, this approach is unsatisfactory because many loss functions (such as MSE and SSIM) are sensitive to pixel shifting. For example, the learned patterns are shifted slightly upward in the learned pattern with RIP loss of Fig. \ref{Fig_VisualLearn}, to capture the forehead, eye, and nose.

\subsubsection{Similarity Loss.} 
\begin{figure*}[!t]
	\centering		
	\includegraphics[width=1\textwidth]{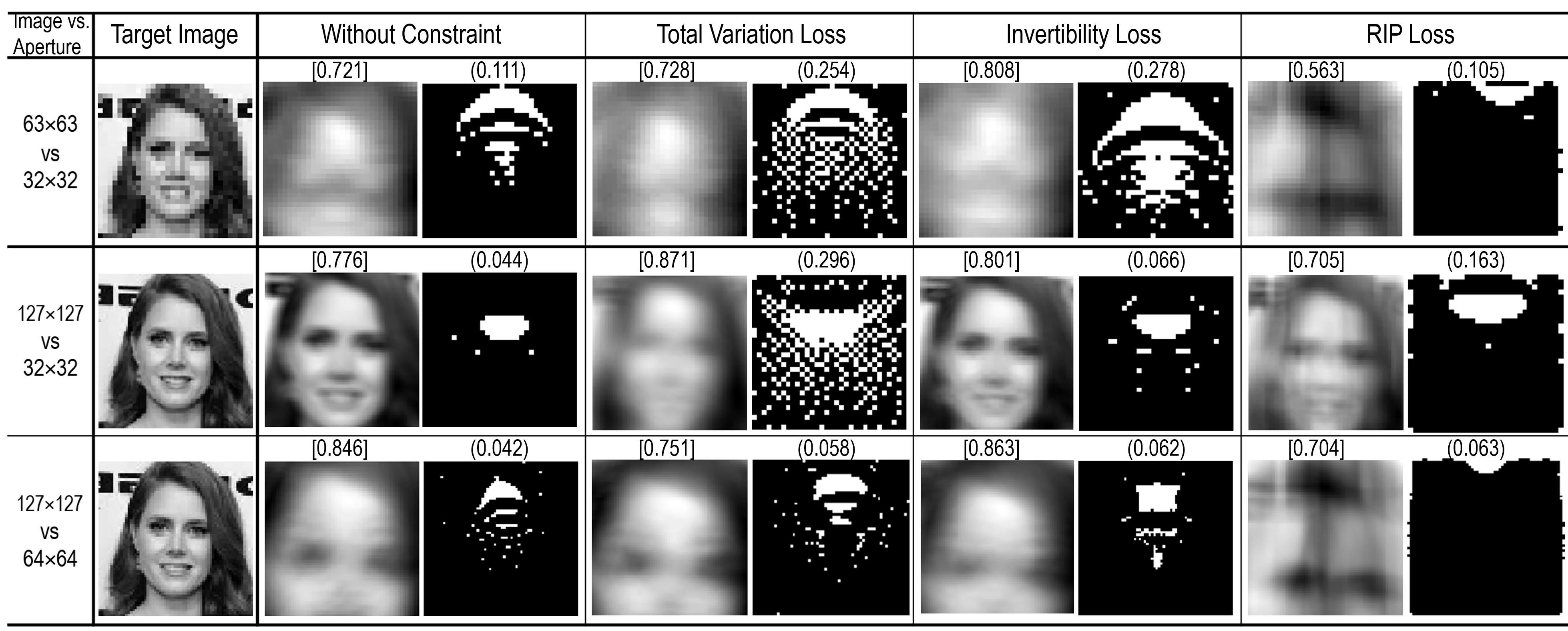}
	\caption{Visual comparison across examples of learned lensless imaging for various combinations of image size ($63\times63, 127\times127$) and pattern size ($32\times32, 64\times64$). Numerical figures shown above measurement images and corresponding patterns indicate [blurriness] $\uparrow$ \cite{crete2007blur} and (aperture ratio) $\uparrow$, respectively. Ranking in terms of decreasing revealed privacy is: fully open, random patterns, learned patterns with constraints, and learned patterns without constraint.}
	\label{Fig_VisualLearn}
\end{figure*}

To overcome the above drawbacks, we propose an alternative approach that uses images captured with a fully open aperture as the most visually secure images (examples are shown in the measured image of the Full open pattern in Fig. \ref{Fig_VisualLearn}). Instead of maximizing the blurriness of captured images, we minimize the difference between the learned measurement and the images captured with a fully open aperture. We, therefore, regard the fully open aperture as the most visually secure image, and we minimize:
\begin{equation}
\mathcal{L}_{\mathsf{sim}} =  \sum_{i} || H * x_i -  {1}_m  * x_i||^2_2,
\end{equation}
where ${1}_m$ denotes the all-ones matrix, which is the coded pattern for fully open aperture imaging, and $x_i$ denotes the $i^{\textrm{th}}$ image in the training data. We use mean squared error for simplicity here, but other loss functions (such as L1 norm) or perceptual metrics (such as SSIM) could also be adopted.

\subsection{Pattern-Based Human-Imperceptible Loss}
This section presents various loss functions $\mathcal{L}_{\mathsf{hi}}$ as constraints to learn the coded $H$ pattern.

\subsubsection{Total Variation Loss}

As shown in all learned patterns without constraint Fig. \ref{Fig_VisualLearn}, in the absence of constraints the learned patterns converged to a local or pinhole-like pattern. As a result, measurements were convoluted from a small region of the images, thus revealing visual information. Unfortunately, similarity loss could not overcome this drawback. However, as the pattern converges locally, its TV also becomes small. A high TV pattern combines distant pixels and reduces local correlations. Therefore, the TV \cite{canh_tv} of $H$ is maximized with:
\begin{equation}
\mathcal{L}_{\mathsf{tv}} = - ||\Delta_x H  ||_1  -|| \Delta_y H ||_1,
\end{equation}
where $\Delta_x$ and $\Delta_y$ represent horizontal and vertical gradient operators, respectively. With TV loss, as can be seen from Fig. \ref{Fig_VisualLearn} the learned patterns become more diverse and degrade visual information.

\subsubsection{Invertibility Loss}

\begin{table*}[!t]
	\centering
		\renewcommand{\arraystretch}{1.5}	
	\caption{Top 1 Accuracy (\%) $\uparrow$ of various imaging setups with ResNet18.}
	\label{tab_acc}
	\begin{tabular}{l|l|c|c|c|c|c|c|c|c|c}
		\hline \hline 
		\multirow{4}{*}{Patt.} & Dataset      & \multicolumn{3}{c|}{MS-Celeb}           & \multicolumn{3}{c|}{VGG-Face2}          & \multicolumn{3}{c}{CASIA}             \\ \cline{2-11} 
		& Image size   & 63$\times$63    & \multicolumn{2}{c|}{127$\times$127} & 63$\times$63    & \multicolumn{2}{c|}{127$\times$127} & 63$\times$63   & \multicolumn{2}{c}{127$\times$127} \\ \cline{2-11} 
		& Patt. size & 32$\times$32    & 32$\times$32         & 64$\times$64        & 32$\times$32       & 32$\times$32            & 64$\times$64           & 32$\times$32      & 32$\times$32            & 64$\times$64     \\ \cline{2-11} 
		& Coded ratio & 1/4    & 1/16         & 1/4        & 1/4       & 1/16            & 1/4           & 1/4      & 1/16            & 1/4           \\ 
		\hline \hline
		\multirow{4}{*}{Fix}     & Pinhole     & \textcolor{red}{\textbf{98.57}}    & {98.51}         & \textcolor{red}{\textbf{99.35}}  & \textcolor{red}{\textbf{96.39}} & \textcolor{red}{\textbf{97.68}} & \textcolor{red}{\textbf{98.89}} & \textcolor{red}{\textbf{95.10}} & \textcolor{red}{\textbf{97.40}} & \textcolor{red}{\textbf{96.20}}        \\ \cline{2-11} 
		& Full-open      & 76.88    & 92.74       & 77.96        & 48.21    & 74.36         & 76.89        & 64.01   & 83.46         & 65.16        \\ \cline{2-11} 
		& Random  & 74.77    & 92.58         & 78.52        & 50.36    & 75.42         & 82.90        & 66.05   & 85.27         & 69.24        \\ \hline
		\multirow{5}{*}{Learn}   
		& LwoC   & 92.36    & \textcolor{red}{\textbf{99.17}} & 95.77 & \textbf{87.67} & \textbf{95.00} & 88.92 & 89.19 & \textbf{95.00} & \textbf{92.75}  \\ \cline{2-11} 
		& LwC-Sim      & 93.12    & 99.02         & 91.11        & 86.54    & 94.18         & 87.11        & 89.31   & 94.23        & 90.89        \\ \cline{2-11} 
		& LwC-TV       & \textbf{93.23}    & \textbf{99.04}         & \textbf{96.18}        & 86.37    & 94.16         & \textbf{91.63}        & \textbf{89.77}   & 94.10         & 90.38        \\ \cline{2-11} 
		& LwC-Inv      & 85.78    & 92.38         & 84.54        & 75.60    & 93.49         &73.44         & 84.51   &    91.47           & 83.23             \\ \cline{2-11} 
		& LwC-RIP      & 91.12    & 95.18         & 93.20        & 78.62    & 89.53         & 81.31        & 88.80   & 92.67         & 86.68             \\ 
		\hline \hline 
		\multicolumn{11}{r}{\scriptsize Best results are in \textcolor{red}{\textbf{red bold}}, second best in \textbf{bold}.}
	\end{tabular}
\end{table*}

Even though it is difficult to reconstruct an image from a lensless measurement because of its poor invertibility \cite{FlatCam,Canh_2019_ICCV,canh_tci21}, it is still possible to do so. Therefore, visual privacy protection is not fully preserved. Unfortunately, there has been limited research on invertibility for lensless imaging. Yang et al. \cite{Yang_2021} proposed an invertibility regularization algorithm to improve reconstruction performance. Regularization biases the learned pattern towards a pinhole pattern. For a binary pattern $H$, invertibility loss is simplified by maximizing the L1-norm of the learned pattern. Unlike \cite{Yang_2021}, we minimize the potential for reconstruction, and define loss as the negative function:
\begin{equation}
\mathcal{L}_{\mathsf{inv}} =  -|| H||_1.
\end{equation}
Minimizing $\mathcal{L}_{\mathsf{inv}}$ leads to a pattern with a large aperture opening. The learned pattern makes the captured image blurry, and difficult to reconstruct.

\subsubsection{Restricted Isometry Property Loss} 
One obvious break in the visual privacy of lensless imaging is to reconstruct the input image from lensless measurement. Therefore, visual privacy is directly related to the ability to reconstruct lensless measurements. In this regard, restricted isometry property (RIP) \cite{candes_rip} is a well-known condition for guaranteeing the reconstruction of sparse signals at a high probability. As in \cite{candes_rip}, RIP condition is formalized as
\begin{equation}
\label{rip_eq}
(1-\delta) ||x ||_2^2 \leq ||H * x||_2^2 \leq (1+\delta) ||x||^2_2,
\end{equation}
The expression above states that ``if a pattern $H$ preserves distance at small $\delta$ for all $s$- sparse signals $x$, then there is a high probability that $x$ can be reconstructed with high quality".
The better in satisfying the RIP condition of measurement, the higher chance to reconstruct the original signal at high quality. The inverse is also true as, the less satisfying the RIP condition is, the less guarantee of reconstruction of the signal. In our goal, we would like to capture lensless measurements with poor RIP properties to make it more difficult to reconstruct. Therefore, we would like to minimize the RIP condition. 


In order to do so, we simplified RIP condition by dividing all side in Eq. \ref{rip_eq} by the signal energy as 
\begin{equation}
\label{rip_eq2}
(1-\delta)  \leq \frac{|| H*x||_2^2 }{|| x||_2^2 + \epsilon} \leq (1+\delta),
\end{equation}
where $\epsilon = 10^{-10}$ is a small constraint added to avoid division by zero. 
As the learned pattern $H$ in lensless imaging is a binary matrix, measurement energy is always smaller than signal energy $|| H*x||_2^2 \leq ||x||_2^2$. Therefore, only left equation of Eq. \ref{rip_eq2} is necessary as 
\begin{equation}
\label{rip_eq3}
(1-\delta)  \leq \frac{|| H*x||_2^2 }{|| x||_2^2 + \epsilon}.
\end{equation}

From Eq. \ref{rip_eq3}, minimizing RIP condition of lensless imaging is equivalent to maximize the isometric constraint $\delta$, or minimizing the ratio between energy of measurement and signal. 
Unfortunately, the RIP condition is nonlinear because it must be satisfied for all $s$-sparse signals, thus, it is difficult to utilize RIP in a learning framework. 
We, therefore, approximate a small RIP condition for only $n$ sample in each training batch by maximizing the negative energy ratio as
\begin{equation}
\label{rip_loss_eq}
\mathcal{L}_{\mathsf{rip}} = - \sum_{i} \frac{|| H*x_i||_2^2 }{|| x_i||_2^2 + \epsilon}.
\end{equation}

By doing so, $\mathcal{L}_{\mathsf{rip}}$ can be calculated and integrated to the neural training framework as a loss function. 

It is also worth mentioning that, the curve of isometric constraint is used to evaluate the visual privacy performance in Section V.


\section{Experimental Results}
\label{sec:exp}
\subsection{Dataset and Training}
In evaluating the proposed human-imperceptible recognition system, we set up a facial identification system with various fields of view from three datasets: VGG-Face2 \cite{cao2018vggface2}, aligned Microsoft Celeb (MS-Celeb) \cite{guo2016ms}, and CASIA \cite{yi2014learning}. For each dataset, we selected the ten classes containing the largest number of images, and then divided them into training and test sets with a ratio of 95:5. Images were converted to gray-scale and resized to the three different specifications $\{n,m\}=\{63\times 63, 32\times 32 \}, \{127\times 127, 32\times 32 \}$, and $\{127\times 127, 64\times 64 \}$, where $n, m$ indicate pattern size of the captured image and coded pattern. For training augmentation, we used random cropping with a padding size of 4 and vertical flipping (with approximately 6520 images per epoch).

Each network was associated with a loss function defined for a fixed or learned coded pattern. Networks were trained from scratch with a stochastic gradient descent optimizer. The initial learning rate was 0.2 and was reduced to a fifth of its value every 100 epochs. We used 600 epochs for training, a momentum of 0.9, and a weight decay of $5\times 10^{-4}$. The mini-batch size was 128. The epoch with the best top-1 classification accuracy was selected as the final solution. In the simulation results, we independently trained each dataset.

\begin{figure*}[!t] 	
	\centering
	\includegraphics[width=1\textwidth]{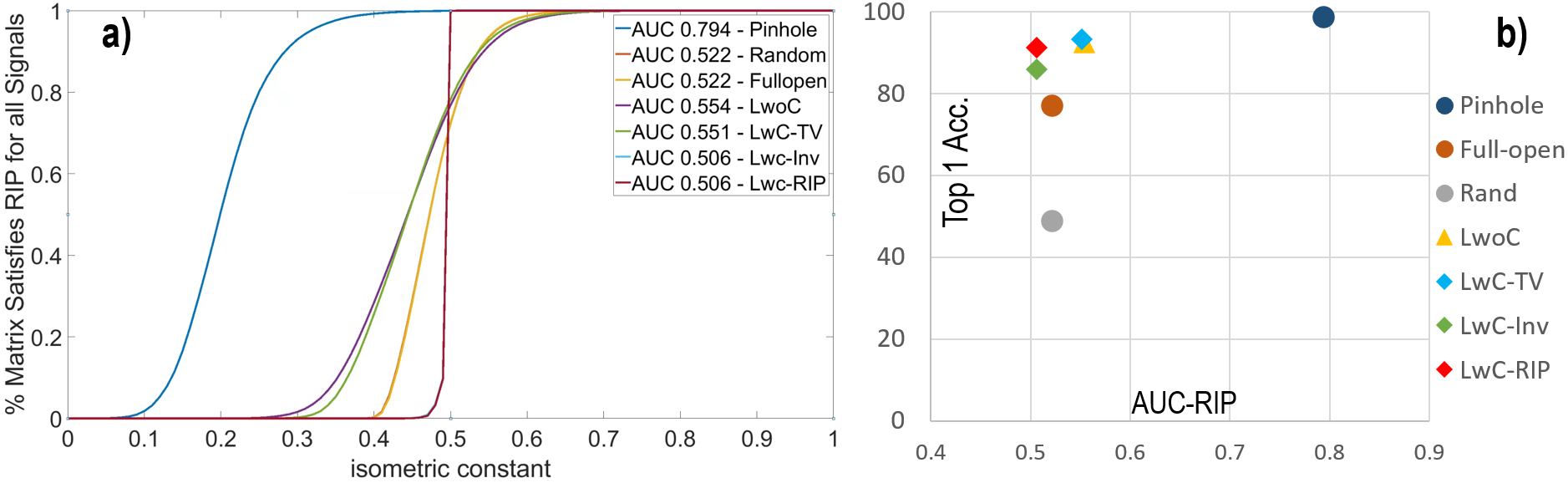}
	\caption{(a) Invertibility performance with AUC-RIP $\downarrow$ and (b) tradeoff between accuracy and RIP. Random and full open patterns achieved the same AUC-RIP. Learned pattern with constraints LwC-Inv and LwC-RIP also shared the same AUC-RIP, but LwC-RIP is superior because of its higher accuracy.}
	\label{fig_rip}
\end{figure*}

\subsection{Benchmark Methods}

We evaluated lensless imaging with both fixed and learned patterns. For lensless imaging with fixed patterns, we tested four identification systems that used a pinhole pattern (indicated by Pinhole), a fully open pattern (Full-open), and a random pattern. For lensless imaging with learned patterns, different methods were trained without privacy constraints as LwoC. Finally, methods that relied on human-imperceptible loss are named LwC-Sim, LwC-TV, LwC-Inv, and LwC-RIP.

\subsection{Identification Results on Simulated Data}
\label{Sec_exp_on_simulation_data}
It is challenging to evaluate human recognition of images because there is relatively little quantitative research on measuring the ability of the human visual system  to recognize objects. In general, it is more difficult for humans to determine subject identity from images when they are blurred. Therefore, we adopted a no-reference objective blur metric \cite{crete2007blur} to evaluate the quality of visual privacy.
Table \ref{tab_acc} shows that the highest accuracy was obtained for pinhole imaging. Lensless imaging with fully open and random patterns resulted in a 20\%--40\% accuracy loss. When evaluated using human-imperceptible identification (shown in the measured image of the Pinhole pattern in Fig. \ref{Fig_VisualLearn}), pinhole imaging revealed image details, while imaging with fully open and random patterns did not. As expected, the learned patterns improved recognition accuracy, yielding approximately a 5\% loss in accuracy compared with pinhole imaging, but failed to guarantee visually protected measurements. As we expected, without any constraint, the learned pattern in LwoC-woRec achieved the second-best recognition performance. Unfortunately, it revealed more visual information. For instance, at a small coded ratio $r=n/m$, LwoC-woRec revealed subject identity as shown in all measured images Without Constraint in Fig. \ref{Fig_VisualLearn}. Therefore, it is desirable to develop a method to control the tradeoff between accuracy and privacy.

\subsubsection{Learned patterns and loss functions}

The recognition accuracy of the constrained methods decreases in the following order: LwC-Sim, LwC-TV, LwC-RIP, and LwC-Inv. However, similarity loss is ineffective in protecting visual privacy. Additionally, blurriness is not a good quality index. For example, Fig. \ref{Fig_VisualLearn} demonstrates that the degree of blurriness estimated via invertibility loss (rows 2--4, column 7) was greater than that estimated via RIP loss (rows 2--4, column 9), however, the corresponding image reveals more visual information. 

To better assess invertibility in a perceptually meaningful manner, we used RIP as a quality metric. First, the RIP of matrix $H$ is represented as a RIP curve for a given set of data. In a RIP curve, at a given $\delta \in [0, 1]$, we measure the percentage of pattern $H$ that satisfies the RIP condition \cite{canh_rsrm}. To obtain an objective score, we calculate the area under the curve (AUC) of the RIP curve, which is named AUC-RIP (see Fig. \ref{fig_rip}-a). The more difficult it is to reconstruct the image from the captured measurement, the better visual protection offered by pattern $H$ against reconstruction. As a rule of thumb, a smaller AUC-RIP is preferred. LwoC and LwC-TV returned higher AUC-RIP scores, indicating that they were easier to reconstruct than the random or fully open patterns. LwC-RIP and LwC-Inv are the top two methods, with significantly smaller AUC-RIP values than the random and fully open patterns. With respect to the trade-off between invertibility and recognition accuracy, our proposed LwC-RIP outperforms all methods considered.

\begin{figure*}[!t] 	
	\centering
	\includegraphics[width=1\textwidth]{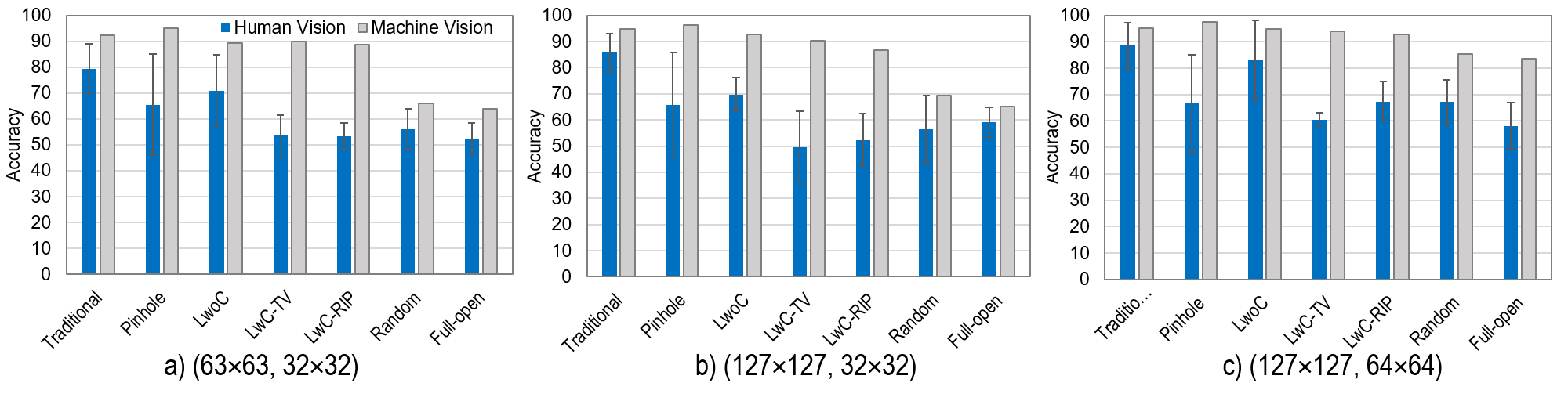}
	\caption{Subjective verification performances $\downarrow$ at various (image, pattern) resolutions. The identification accuracies $\uparrow$ were taken from Table~\ref{tab_acc} for references.}
	\label{fig_subjective_accuracies}
\end{figure*}

\subsection{Subjective Evaluation on Simulated Data} \label{sec:subjectTest}

\subsubsection{Experimental Setup}
In this perceptual test, we pre-selected a dataset of frontal faces by retaining the first 1200 identities from the CASIA dataset. For each identity, we selected the three frontal face images with the highest quality. We generated corresponding simulated lensless images for face verification by convolving them with the patterns detailed in Section~\ref{Sec_exp_on_simulation_data}. We adopted three configurations for image and pattern size: $(n,m) \in \{(63\times63,32\times32),(127\times127,32\times32),(127\times127,64\times64)\}$.
We showed a pair of facial images, with the same identity or different identities, to human observers, and asked them to judge whether the two images featured the same subject. We recruited ten observers without prior knowledge of the identity of the test faces. The ratio between identical and different identity pairs was 50:50. We used VQOne~\cite{vqone} for the perceptual tests. Each volunteer labeled a total of 560 pairs of facial images (180 pairs for each size configuration, which is equivalent to 30 pairs for each coded pattern). Pairs of facial images for the same coded patterns were shown randomly, to avoid the possibility that observers may have explicit knowledge about which $H$ pattern was used. In terms of test condition, the test duration was between 30 and 60 minutes, and matching accuracy was used for verification results.

\begin{table}[!t]
	\centering 
	\renewcommand{\arraystretch}{1.2}	
	\caption{Top-1 Accuracy (\%) with real measurement from selected CASIA dataset. }
	\label{Tab_RealAcc}
	\small
	\setlength{\tabcolsep}{3pt}	
	
	\begin{tabular}{|p{3cm}||c|c|c|c|c|}
		\hline \hline 
		Imaging	                &\multicolumn{2}{c|}{ResNet18 $\uparrow$ }  & \multicolumn{2}{c|}{Blurriness*  $\uparrow$} \\ \hline \hline 
		Measurement size        & 64$\times$64 & 512$\times$512 & 64$\times$64 & 512$\times$512 \\ \hline
		Pinhole                 & 89.17 & 89.33 & 0.44 & 0.14  \\ \hline
		Full-open               & 46.33 & 45.67 & 0.50 & 0.19  \\ \hline
 		Rand                    & 55.17 & 54.14 & 0.77 & 0.32  \\ \hline
		LwoC                    & 88.00 & 87.67 & 0.82 & 0.36  \\ \hline
		LwC-TV                  & 87.00 & 85.50 & 0.80 & 0.40  \\ \hline 
		LwC-RIP                 & 82.50 & 80.33 & 0.80 & 0.42  \\ \hline \hline 
		\multicolumn{5}{r}{\scriptsize *Contrast enhancement impacts blurriness, especially for full-open.} 
	\end{tabular}
\end{table}

\subsubsection{Subjective Results for Face Verification}

Fig.~\ref{fig_subjective_accuracies} shows that the subjective verification accuracy for pinhole imaging was highest, while accuracy values for random and fully open patterns were lowest because details were blurred in the associated images. Verification accuracy values for LwC-TV and LwC-RIP were similar to the values obtained for random and fully open patterns. In addition, accuracy values for the learned patterns increased when human-imperceptible constraints were removed (compare LwoC with LwC-TV and Lwc-RIP) .

The verification results demonstrate that privacy protection constraints (LwC-TV and LwC-RIP) could protect privacy as effectively as random or fully open patterns. Their machine vision accuracy values for identification were also similar to those associated with pinhole imaging. These experiments confirm that our proposed losses achieved both visual privacy and high recognition accuracy. On the other hand, lower image resolution and aperture resolution ratios lead to improving privacy protection.

\subsection{Identification Results on Real Imaging Systems}
\begin{figure*}[t] 	
	\centering
	\includegraphics[width=1\textwidth]{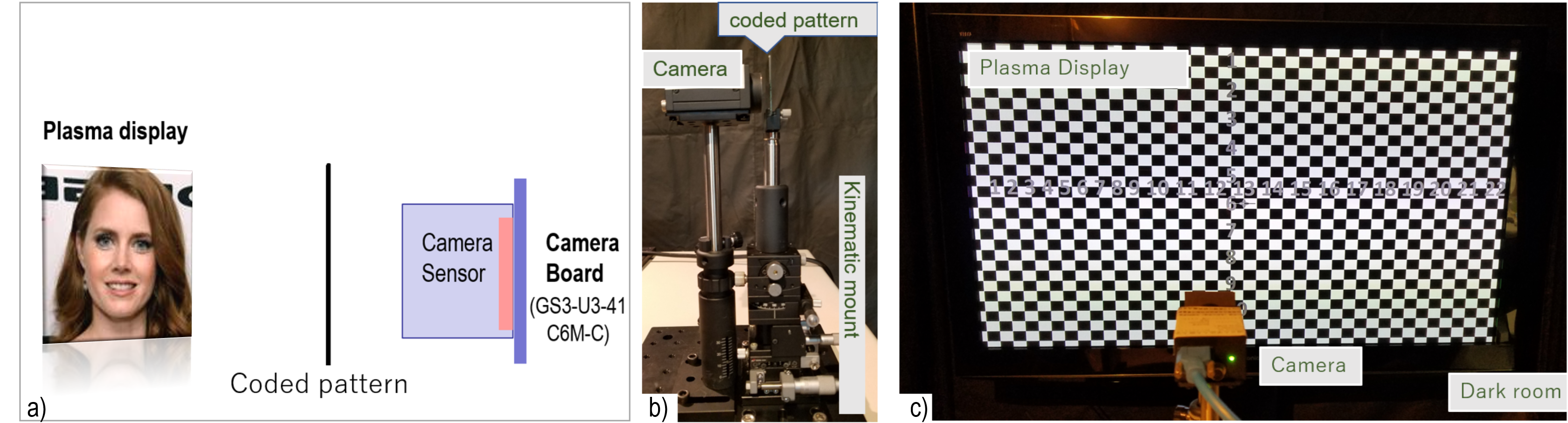}
	\caption{ a) Diagram of the experimental setup, b) the sensor alongside a photolithography mask on a 5D kinematic mount and c) the optics system in front of a plasma display in a dark room. }
	\label{fig_hardware_printed}
\end{figure*}

\begin{figure*}[!t] 	
	\centering
	\includegraphics[width=1\textwidth]{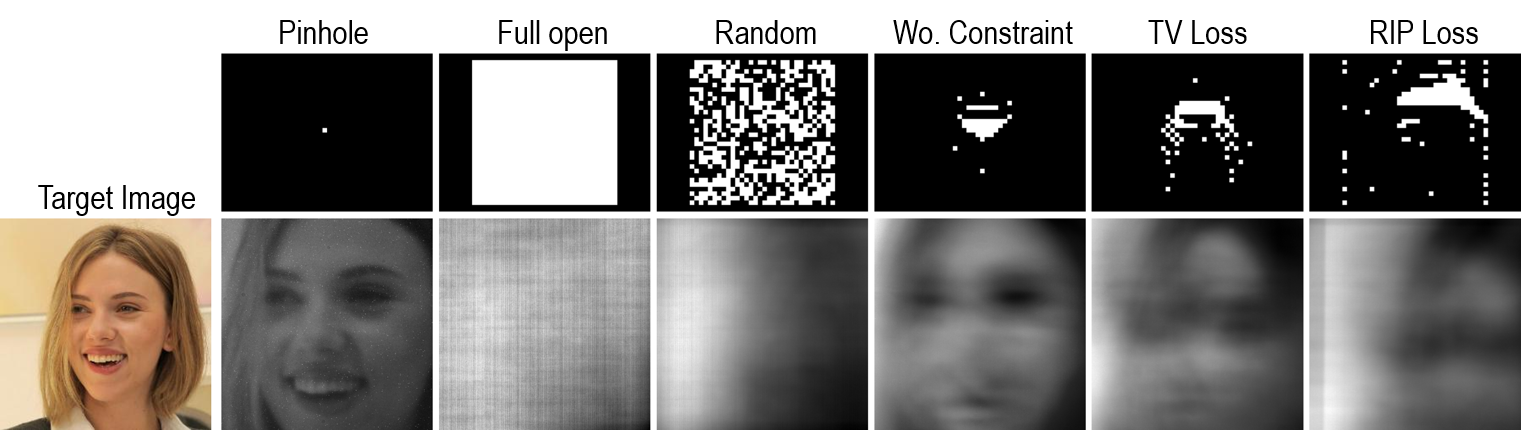}
	\caption{Realcaptured measurements with various printed patterns: pinhole, full open, learned without constraint (LwoC), learned with TV loss (LwC-TV), and learned with RIP loss (LwC-RIP). }
	\label{RealCaptured_printed}
\end{figure*}

We selected the smallest resolution setting (i.e., $\{63\times 63, 32\times 32 \}$) from Table~\ref{tab_acc}, and implemented a prototype system with printed patterns at $d_1$ = $22$mm. A plasma monitor for displaying images was placed at a distance of $80$cm from the printed patterns in a dark room. The coded patterns were printed with chromium on glass using photolithography printing. The area of the mask measured $5.6\times5.6$ mm. Fig.~\ref{fig_hardware_printed} shows the system prototype and the experimental setup.
We chose the ten CASIA classes with the most images and included the largest images (in terms of pixel size) from each class in our dataset. The captured images were normalized adaptively to increase their contrast, cropped to the same size as the mask, and resized to measure $b \times b$ pixels. We experimented with $b\times b$ = 512$\times$512 and 64$\times$64 pixels. For each identity, 310 and 60 images were captured for training and testing, respectively, as shown in Fig.~\ref{RealCaptured_printed}.

In real imaging scenarios, image quality is generally poorer than the quality associated with simulated experiments. Nevertheless, images captured at low resolution with the pinhole pattern are still relatively sharp, as shown in figure \ref{RealCaptured_printed}. For other coded patterns, the amount of residual visual information is in line with the simulations and almost invisible to human vision, particularly for fully open and random patterns. These results demonstrate that lensless images successfully prevent human vision from accessing information about identity. In line with the simulations, we did not record any substantial privacy information from measurements with fully open and random patterns (at 50\% aperture ratio). Without constraints, the learned pattern without constraint (LwoC) revealed more information. For the face recognition application, because of the inevitable difference between real and simulated measurements, we re-trained a ResNet18 network to handle noise and low-contrast images. Table \ref{Tab_RealAcc} reports the resulting of top-1 accuracy values achieved on real test data.

Despite its high performance in simulations, pinhole imaging performed poorly with a real dataset because of inefficient light capture at small coded ratios. In addition, the captured images had very low contrast because they were highly blurred. Notwithstanding these shortcomings, pinhole images still achieved the highest recognition accuracy (89.17\%). However, similarly to the simulated results, subject identity could be retrieved from these images by the human visual system. In contrast, fully open imaging provided no visual information but produced the worst performance in the recognition task, with only 46.33\% accuracy. Random patterns yielded slightly better performance at 55.17\%, but these values are still substantially smaller (by approximately 28\%--33\%) than corresponding values obtained with learned patterns without constraints (LwoC) and learned pattern with constraint (i.e. TV loss/LwC-TV, RIP loss/LwC-RIP). LwC-TV performed better than LwC-RIP with accuracy values of 87.00\% and 82.50\%, respectively, but it also reveals more information as shown in Fig. \ref{RealCaptured_printed}.
In general, performance on real data was consistent with the results of our simulations, except for an overall reduction in accuracy of 1.2\%--5\%. The smaller measurement size (64$\times$64) produced more blurriness and less noise than the larger measurement size (512$\times$512). As a result, we observed a small reduction in accuracy for Pinhole and a small increase in accuracy for other patterns.

Although the performance on real data may need to be further improved for practical applications, this current work has taken a first step in the direction of developing effective privacy-preserving imaging. Furthermore, we have successfully implemented a proof-of-concept for our approach in the form of a physical prototype. In future work, recognition accuracy may be improved by (1) adopting a larger dataset, (2) developing a better hardware implementation, and (3) trading visual privacy for higher accuracy.

\section{Conclusion}
\label{sec:conclusion}
This paper proposes a lensless imaging system to protect visual privacy against perceptual attacks by the human visual system. We achieved this goal via measurements with similarity loss and or patterns with TV, invertibility, and RIP losses. 
Experimental results for both simulation and the hardware prototype and subjective study have demonstrated that the proposed methods are able to balance between human-imperceptibility and recognition accuracy for lensless imaging. We believe this is the first work to construct a learnable hardware-based recognition system that considers visual privacy protection. This paper provides preliminary results that outline challenges in combining strong privacy protection with high recognition accuracy. 



\bibliographystyle{splncs}
\bibliography{egbib}

\end{document}